\documentclass[10pt,twocolumn,letterpaper]{article}

\usepackage[pagenumbers]{cvpr} % To force page numbers, e.g. for an arXiv version

\usepackage{graphicx}
\usepackage{amsmath}
\usepackage{amssymb}
\usepackage{booktabs}
\usepackage{makecell}
\usepackage{comment} 
\usepackage{multirow}

\usepackage[pagebackref,breaklinks,colorlinks]{hyperref}

\usepackage[capitalize]{cleveref}
\crefname{section}{Sec.}{Secs.}
\Crefname{section}{Section}{Sections}
\Crefname{table}{Table}{Tables}
\crefname{table}{Tab.}{Tabs.}

\def\cvprPaperID{****} % *** Enter the CVPR Paper ID here
\def\confName{CVPR}
\def\confYear{2023}

\begin{document}

\title{LinK: Linear Kernel for LiDAR-based 3D Perception}

\author{
Tao Lu $^{1}$ \quad
Xiang Ding $^{1}$ \quad
Haisong Liu $^{1}$ \quad
Gangshan Wu $^{1}$ \quad
Limin Wang $^{1,2}~\thanks{Corresponding author.}$\vspace{0.2em}\\
$^{1}$State Key Laboratory for Novel Software Technology, Nanjing University \quad $^{2}$Shanghai AI Lab \vspace{.2em}\\
{\tt\small \{taolu,xding,liuhs\}@smail.nju.edu.cn, \{gswu,lmwang\}@nju.edu.cn}
}

\maketitle

%%%%%%%%% ABSTRACT
\begin{abstract}
Extending the success of 2D Large Kernel to 3D perception is challenging due to: 1. the cubically-increasing overhead in processing 3D data; 2. the optimization difficulties from data scarcity and sparsity. Previous work has taken the first step to scale up the kernel size from $3 \times 3 \times 3$ to $7\times7\times7$ by introducing block-shared weights. However, to reduce the feature variations within a block, it only employs modest block size and fails to achieve larger kernels like the $21\times21\times21$. To address this issue, we propose a new method, called LinK, to achieve a wider-range perception receptive field in a convolution-like manner with two core designs. The first is to replace the static kernel matrix with a linear kernel generator, which adaptively provides weights only for non-empty voxels. The second is to reuse the pre-computed aggregation results in the overlapped blocks to reduce computation complexity. The proposed method successfully enables each voxel to perceive context within a range of $21\times21\times21$. Extensive experiments on two basic perception tasks, 3D object detection and 3D semantic segmentation, demonstrate the effectiveness of our method. Notably, we rank \textcolor[rgb]{1,0,0}{1st} on the public leaderboard of the 3D detection benchmark of nuScenes (LiDAR track), by simply incorporating a LinK-based backbone into the basic detector, CenterPoint. We also boost the strong segmentation baseline's mIoU with {\bf2.7\%} in the SemanticKITTI test set. Code is available at \href{https:\/\/github.com\/MCG-NJU\/LinK}{https://github.com/MCG-NJU/LinK}.
\end{abstract}

%%%%%%%%% BODY TEXT
\section{Introduction} % 1 pages
\label{sec:intro}

\begin{figure}[t]
   \centering
   \includegraphics[width=0.8\linewidth]{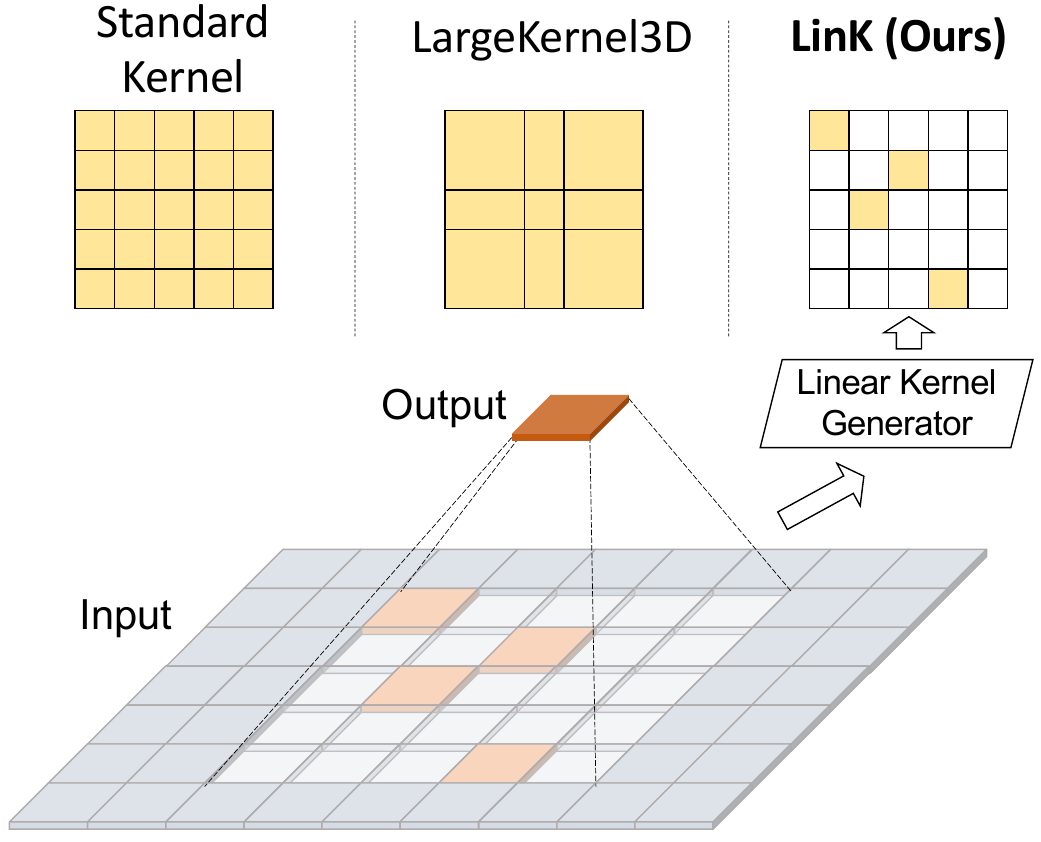}
   	\vspace{-0.2cm}
  \caption{Comparisons among the standard kernel, the LargeKernel3D's~\cite{lk3d} block-shared kernel, and our LinK's kernel from a generator. Instead of storing a dense kernel matrix, LinK generates sparse kernels online according to the input data. The amount of learnable parameters will not increase along with the kernel size, which enables the scaling up of larger kernel.}
  \label{fig:title}
  \vspace{-0.3cm}
\end{figure}

There is a consensus that a large receptive field contributes positively to many downstream vision tasks. For example, Transformer~\cite{dosovitskiy2020image,liu2021swin} benefits a lot from the global relation with self-attention and becomes the leading topic in classification~\cite{dosovitskiy2020image}, segmentation~\cite{xie2021segformer}, and detection~\cite{detr}. However, self-attention is not the only route to a large receptive field. Previous works like the RepLKNet~\cite{replknet} and SLaK~\cite{liu2022more} investigated the potential of obtaining wide-range information through a large convolutional kernel. They have achieved comparable results with the Transformer-based methods. Considering that the convolution operator is more friendly to existing chip architecture, the large kernel method is efficient in real applications. This raises an immediate question in the 3D perception: can the philosophy of large kernel generalize to the 3D task? 

The answer is yes. LargeKernel3D~\cite{lk3d} takes the first step and successfully achieves better metrics on both segmentation and detection. Time and space consumption are core concerns during the extension since they increase cubically in the 3D task. LargeKernel3D~\cite{lk3d} introduces a spatial sharing kernel to scale the 3D kernel up to $7\times7\times7$ and restrict the rapid growth of parameters amount. However, compared with the 2D counterparts, which have developed the huge size of $31\times31$~\cite{replknet} and even $51\times51$~\cite{liu2022more}, the $7\times7\times7$ seems to be not large enough, hence only benefiting from limited context. There are at least two reasons to hinder its size expansion: first, although the parameter amount is under control, the total amounts of operation on each voxel are still increasing cubically; second, its assumption that the outer parts can share a block-wise weight is too strong to work well in a larger block. So, enlarging the 3D kernel size effectively and efficiently is still a challenging problem.

To handle these issues, we propose a new method, called LinK, to implement a wider-range perception in a convolution-like manner. Two core designs make up the method. The first is to replace the static kernel weights with a linear kernel-generating module to provide weights only for those non-empty areas since the 3D input is very sparse. Meanwhile, this module is layer-wisely shared, which avoids the circumstances that some weights allocated to the blank spaces are not optimized in one iteration. The second is to reuse the pre-computed aggregation results in the overlapped blocks, which makes the computation complexity independent of the kernel size. In other words, we can implement arbitrary kernel sizes with consistent overhead based on the proposed LinK. Brief comparisons among the proposed method and other methods are depicted in Fig~\ref{fig:title}. 

Extensive experiments on the public benchmarks of 3D detection and semantic segmentation tasks demonstrate the effectiveness of LinK. Notably, we achieve the \textcolor[rgb]{1,0,0}{1st} place on the famous 3D detection leaderboard, nuScenes (LiDAR track)~\cite{nuscenes}, by simply replacing the backbone of a classic detection method with the LinK-based backbone. As for the segmentation task, we boost the strong baseline's mIoU with {\bf2.7\%} in the SemanticKITTI test split~\cite{semantickitti}. We will unfold the details in the following sections.

\section{Related Work} 
\label{sec:relatedworks}

\subsection{3D Backbone}

According to the input data format (without considering the multi-view 2D representations in this paper), the 3D backbones are grouped into the voxel-based and the point-based method. 

Early voxel-based methods~\cite{maturana2015voxnet} directly adopt the 3D convolutional layers to process the volumetric data at the cost of cubic growth of time-space complexity, which forbids a fine voxel resolution. Some researchers propose to optimize the 3D convolution with a compact data structure. Octree-based methods~\cite{riegler2017octnet,wang2017ocnn} introduce to organize the data into an octree. PVCNN \cite{liu2019pvcnn} chooses to keep the coarse voxel resolution and compensates for the geometric details with a point branch. Other researchers leverage sparse convolutions to reduce the computation overhead. To solve the submanifold dilation problem of regular sparse convolutions, Graham \textit{et al.} \cite{graham20183d} propose submanifold sparse convolutional networks (SSCNs) that keep the same level of sparsity throughout the network. MinkowskiNet \cite{choy20194d} proposes 4-dimensional convolutional neural networks for spatio-temporal perception. All these methods are restricted to small kernel sizes, and they enlarge the receptive field by stacking more layers. Inspired by RepLKNet~\cite{replknet}, LargeKernel3D~\cite{lk3d} explores ways to scale up the small kernel to a large one. 

To exploit the disorder, point-based methods directly learn from the points' coordinates without any voxelization or projection. The series of PointNet~\cite{qi2017pointnet,qi2017pointnet++} learn point-wise features with MLP and aggregate global features with max-pooling. To imitate the 2D convolution, some position-adaptive kernel generation methods~\cite{liu2019relation,thomas2019kpconv,wu2019pointconv,boulch2020convpoint,li2018pointcnn,xu2018spidercnn} are proposed to learn the spatial kernel distribution. Another line of work \cite{simonovsky2017dynamic, wang2019dgcnn} focus on graph-based networks, which consider each point in a point cloud as a vertex of a graph and generate directed edges based on the neighbors of each point. To purse an efficient network, APP-Net~\cite{appnet} proposes a "push-pull" operator to reduce the redundancy in overhead. Recently, encouraged by the success of Transformer \cite{vaswani2017attention} in natural language processing, PCT \cite{guo2021pct} and Point Transformer \cite{zhao2021pt} design self-attention layers for point clouds and construct Transformer networks for various tasks.

\subsection{LiDAR-based 3D Perception}

\paragraph{Detection.} LiDAR-based 3D detection aims to predict 3D rotated bounding boxes of objects from point clouds. Here, we mainly focus on outdoor scenarios. SECOND \cite{yan2018second} improves VoxelNet~\cite{maturana2015voxnet} by optimizing sparse 3D convolutions. PointPillars~\cite{pointpillars} replaces voxel representation with a pillar one, which organizes the point clouds in vertical columns for better efficiency. Some researchers also extend 2D detection frameworks to 3D space and achieve remarkable results. Inspired by the R-CNN family~\cite{fastrcnn,fasterrcnn} for 2D detection, PointRCNN~\cite{pointrcnn} and PV-RCNN~\cite{pvrcnn} use similar detection pipelines with two stages. CenterPoint~\cite{centerpoint} is evolved from CenterNet~\cite{duan2019centernet} and CenterTrack~\cite{zhou2020tracking}, which adopts a center-based representation for 3D objects. Transfusion-L~\cite{transfusion} is a query-based detector (like DETR~\cite{detr} and Deformable-DETR~\cite{zhu2020deformable}) and builds a transformer decoder with a small set of object queries. 

\vspace{-0.5cm}

\paragraph{Segmentation.} In 3D semantic segmentation, one is required to infer the label of each 3D point. The point-based methods~\cite{qi2017pointnet++,randla} face the overhead issue since their random access to memory is too expensive to deal with the large-scale outdoor scene. Voxel-based methods dominate this task using sparse convolution. Following PVCNN~\cite{liu2019pvcnn}, SPVNAS~\cite{spvnas}  proposes sparse point-voxel convolution. To tackle the imbalanced distribution of points, Cylinder3D~\cite{cylinder3d} proposes a cylindrical partition and builds a 3D cylinder convolution network. RPVNet~\cite{rpvnet} utilizes three different representations of points (range-based, point-based, and voxel-based) and fuses them into a unified network. For better training of the network, 2DPASS~\cite{2dpass} and PVKD~\cite{pvkd} enhance the network with distillation strategies.

Unlike these methods focusing on improving the architectures or the training process, this paper proposes a universal backbone for 3D perception tasks.

\begin{figure*}[t]
   \centering
   \includegraphics[width=0.95\linewidth]{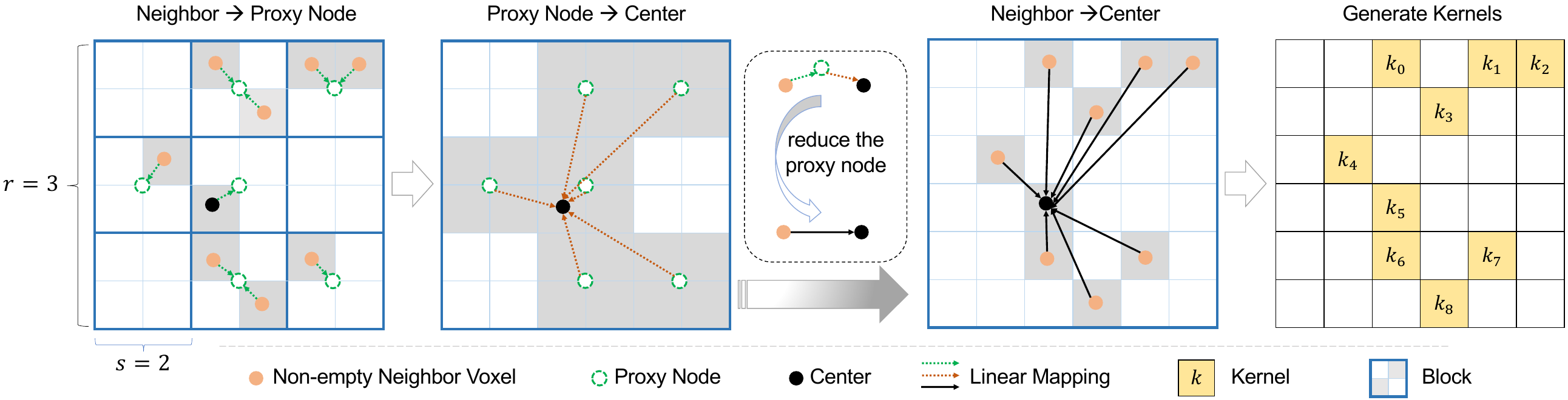}
   \vspace{-0.1cm}
  \caption{The procedure of LinK. The input is partitioned into non-overlapped blocks (the block size is $s^3$). Each non-empty voxel pushes its feature to a block-wise proxy, and then the center voxel pulls features only from the neighbor proxies (neighbor range is $r^3$). The push and pull processes are driven in a reducible manner (details in~\ref{sec:blockagg}) such that the block-wise proxy supports reusing for every potential calling. The resulting matrix serves as a convolutional kernel to weigh neighbors.}
  \label{fig:link_process}
\end{figure*}

\section{Methodology}  
\label{sec:methods}

This section introduces all of the designs of our method. We begin with two backgrounds to clarify our work's innovation in section~\ref{sec:bg}. Then, the detailed procedures are provided in section~\ref{sec:link}. Finally, we show how to incorporate the proposed backbone into two basic 3D perception tasks: object detection and semantic segmentation, in section~\ref{sec:task}. 

\subsection{Background} 
\label{sec:bg}
\subsubsection{Introduction to 3D Sparse Convolution}
\label{sec:bg1}
Convolution-based methods aggregate the weighted influences within a pre-specified range. The weights are determined by the local relative positions to the convolutional center. Formula~\ref{eq:conv} shows the general process of the 3D convolutional operator~\cite{graham20183d,choy20194d}.

\begin{equation}
\label{eq:conv}
    g_p = \sum_{n\in \mathbb{N}}{w_n\cdot f_{p+n}},
\end{equation}
\noindent

\noindent
where $p$ is the convolution center. $\mathbb{N}$ denotes the non-empty neighbors within a specified range. $f_*$ and $g_*$ are the input and output features, respectively. $w_n$ is the kernel corresponding to the relative location $n$. Different from the 2D images, the Lidar data are spatially sparse. This means that kernels allocated to the empty areas will not participate in the convolution computation, leaving them to fail to be updated during backward propagation. This slows down the optimization process. Meanwhile, regardless of the inputs, the kernel of every location must be stored for the potential calling. This causes a cubically-increasing amount of parameters in a large 3D kernel. For example, for a single convolution layer with a kernel size of $21^3$, $C_{in}=32$, $C_{out}=64$, more than 18M learnable parameters are waiting for calling, although most of them will be idle during the inference.

\subsubsection{Introduction to the Push-Pull Strategy}
\label{sec:pp}
The core function of the convolution operation is to introduce spatial interaction. When the convolutional window slides on the feature map, locations covered by the window will be involved in computation with kernels. The overlapped area participates in computation repeatedly, which introduces redundancy. To deal with this issue in the point cloud task, APP-Net~\cite{appnet} proposes a so-called APP operator, which decomposes the spatial interaction into three steps: a push step, $\gamma(p_i\to proxy)$, to push $p_i$'s feature into a cluster-sharing proxy, an aggregation step in the proxy to fuse cluster-wise information, and a pull step, $\lambda(proxy\to p_j)$, to pull feature from the auxiliary proxy for each point $p_j$. Since points within the same cluster share the same aggregation proxy, APP operator avoids the redundant aggregation computation in each cluster. The push step, pull step, and aggregation in the proxy are designed comprehensively to satisfy the requirement that 

\begin{equation}
\label{eq:apprequirement}
	\gamma(p_i\to proxy)\circ\lambda(proxy\to p_j) = \eta(p_i\to p_j).
\end{equation}
\noindent
$\circ$ denotes an operator to combine $\gamma(*)$ and $\lambda(*)$. $\eta(p_i\to p_j)$ is a function to measure the relation between $p_i$ and $p_j$ directly. According to Formula~\ref{eq:apprequirement}, the influence from the proxy is reducible in the final. The design of APP operator is tightly coupled with the point cloud modality, how to activate its ability in processing voxel data is not explored yet.

\subsection{Linear Kernel in 3D} 
\label{sec:link}

According to the statements in section~\ref{sec:bg1}, we conclude that storing the kernel value of every discrete location is neither memory-efficient nor friendly to the optimization process for the 3D large kernel. Thus we propose to adopt a neural network module $w(n)$ to generate the kernel online rather than store the static kernel values $w_n$. This makes the amount of learnable parameters not increase with the kernel size. Furthermore, the empty voxels would not slow down the optimization process.

Although resolving the storing cost and optimization issue, there is still a challenge before adopting the generation module to 3D large kernel. The computations between the kernel and feature map introduce cubically increasing overhead. To deal with it, we provide two critical designs in the following parts: the \texttt{Linear Kernel Generator}, and the \texttt{Block Based Aggregation}. The whole process is depicted in Fig~\ref{fig:link_process}.

\subsubsection{Linear Kernel Generator}
A larger kernel extracts the input information at the cost of processing each area for more times. We devote to finding reusable common parts in the overlapped area to reduce the overhead.

\begin{figure}[t]
   \centering
   \includegraphics[width=1.0\linewidth]{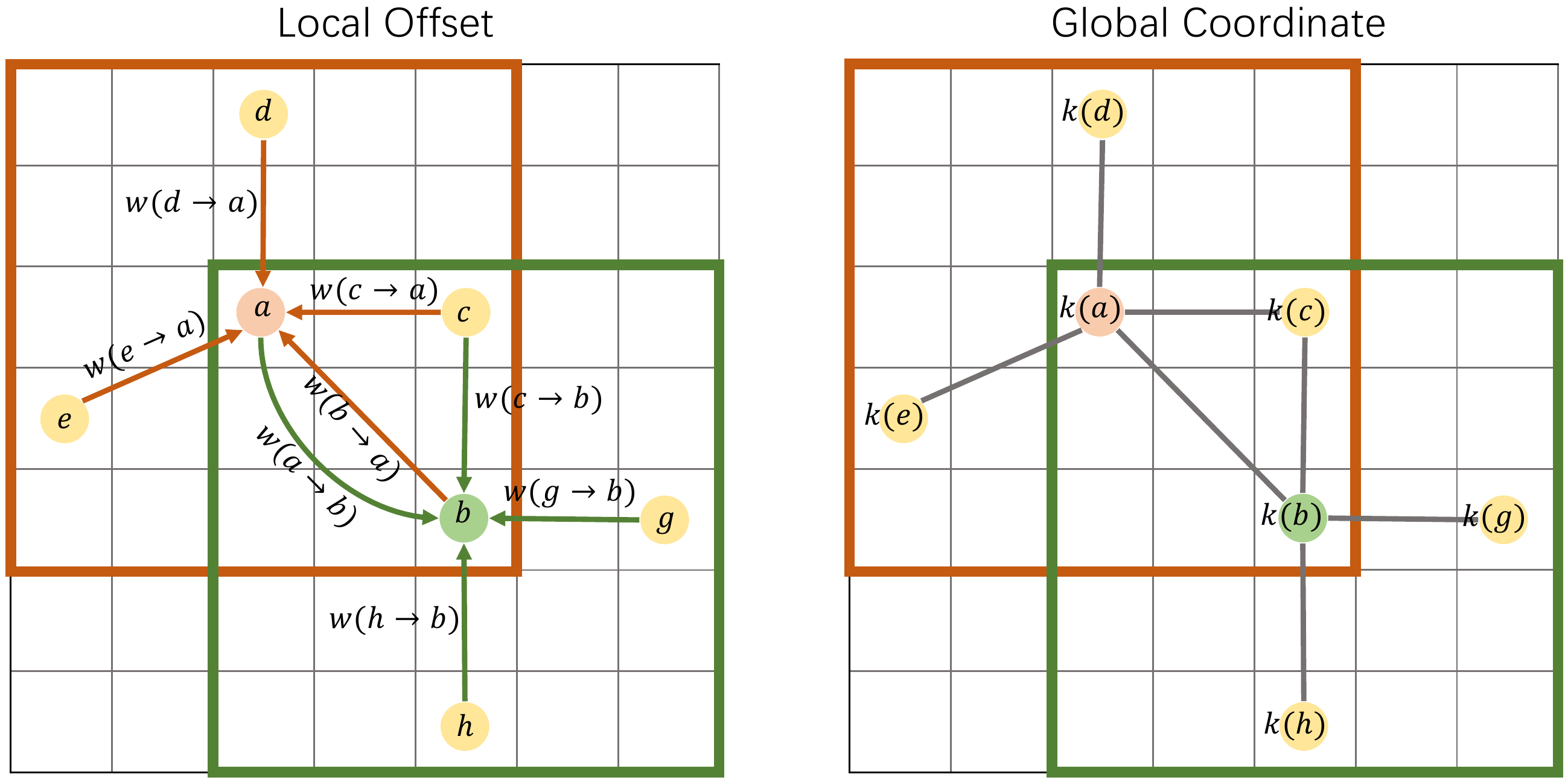}
\vspace{-0.3cm}
  \caption{The comparison between modeling the local offsets and global coordinates. $w(x-y)$ actively measures how $x$ influences $y$ through modeling the offset from $x$ to $y$. $k(x)$ generates kernel for $x$ using its global coordinates. When $y$ queries neighbor features, the relation between $x$ and $y$ can be composed from $k(x)$ and $k(y)$.}
  \label{fig:localglobal}
  \vspace{-0.3cm}
\end{figure}
 
 We start the analysis through a toy example. Given two blocks $A=\{a,b,c,d,e\}$ and $B=\{a,b,c,g,h\}$, where each element denotes a voxel, the overlapped area is $O=\{a,b,c\}, $as shown in Fig~\ref{fig:localglobal}. We try to aggregate $A$'s features to $a$ and $B$'s features to $b$ using the local offsets, and the influences from the overlapped parts are as follows:
\vspace{-0.2cm}
\begin{equation}
\label{eq:local}
	\begin{split}
		\left \{
			\begin{array}{ll}
			    f_{O\to a} = w(a-a)f_a+w(b-a)f_b+w(c-a)f_c,  \\
			    f_{O\to b} = w(a-b)f_a+w(b-b)f_b+w(c-b)f_c.
			\end{array}
		\right.
	\end{split}
\end{equation}

\noindent
According to Formula~\ref{eq:local}, each element in $O$ contributes to $a$ and $b$ using different offsets, so we cannot reuse the overlapped aggregation results through modeling the local offsets.

Considering that the global coordinate for each voxel is fixed and unique, we think about decomposing the local offset into combinations of the global coordinates. Specifically, we first define a new kernel generator as follows:

\begin{equation}
    k(x) = \Psi(\sigma(x))
\end{equation}
\noindent
where $\sigma(x) = W\times x$ is a linear mapping function, $W\in \mathbb{R}^{C_{in}\times 3}$. $\Psi(*)$ is an activation function. Inspired by APP-Net~\cite{appnet}, we adopt trigonometric functions for the activation, e.g., 
\begin{equation}
    \{k^{(0)}(x)=\cos(\sigma(x)),  \hspace{0.5cm}
    k^{(1)}(x)=\sin(\sigma(x))\},
\end{equation}

\noindent
which provides the following relation based on the sum-to-product formula:

\vspace{-0.3cm}
\begin{align}
\label{eq:cos}
    k^{(0)}(x-y) =&\cos(\sigma(x))\cdot \cos(\sigma(y)) \notag \\&+ \sin(\sigma(x))\cdot \sin(\sigma(y)) \notag \\
    =& k^{(0)}(x)\cdot k^{(0)}(y)+k^{(1)}(x)\cdot k^{(1)}(y).
\end{align}

\noindent
where $x$ and $y$ are the global coordinates, and $x-y$ is local offset. Equation~\ref{eq:cos} decomposes local offset into global positions.

Then we compute the following two  auxiliary aggregations for the overlap area:

\vspace{-0.2cm}

\begin{equation}
\label{eq:overlap}
	\begin{split}
		\left \{
			\begin{array}{ll}
			    f^{(0)}_{O} = k^{(0)}(a)\cdot f_a+k^{(0)}(b)\cdot f_b+k^{(0)}(c)\cdot f_c,  \\
			    f^{(1)}_{O} = k^{(1)}(a)\cdot f_a+k^{(1)}(b)\cdot f_b+k^{(1)}(c)\cdot f_c.
			\end{array}
		\right.
	\end{split}
\end{equation}
\noindent
$f^{(0)}_{O}$ and $f^{(1)}_{O}$ are reusable between $A$ and $B$ since they are computed based on the fixed global locations. To obtain the final aggregations of the overlapped area for the center voxel $a$, we leverage the auxiliary aggregations in the following manner:

\vspace{-0.2cm}
\begin{equation} 
\label{eq:lineargenerator}
	\begin{split}
		f_{O\to a} =& f^{(0)}_{O}\cdot k^{(0)}(a)+f^{(1)}_{O}\cdot k^{(1)}(a)\\
			=& \sum_{p\in O}{k^{(0)}(p-a)}\cdot f_{a+(p-a)}.
	\end{split}
\end{equation}
\noindent
Combined Formula~\ref{eq:overlap} with Formula~\ref{eq:cos}, the $f_{O\to a}$ models the local offsets. And the Formula~\ref{eq:lineargenerator} is an instantiation of the convolutional operator in Formula~\ref{eq:conv}. We call this process \texttt{Linear Kernel Generator} to emphasize the linear mapping in the core part.

\begin{figure*}[htb]

\begin{minipage}[b]{0.99\linewidth}
  \centering
  \centerline{\includegraphics[width=0.95\linewidth]{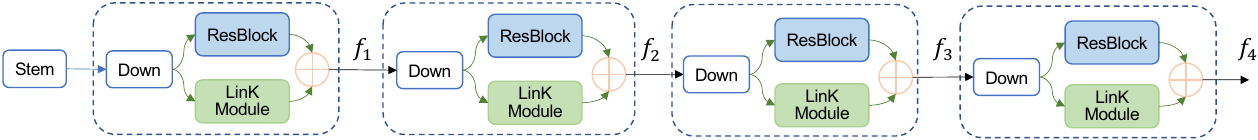}}
  \centerline{(a) Backbone.}\medskip
\end{minipage}
\begin{minipage}[b]{.55\linewidth}
  \centering
  \centerline{\includegraphics[width=0.9\linewidth]{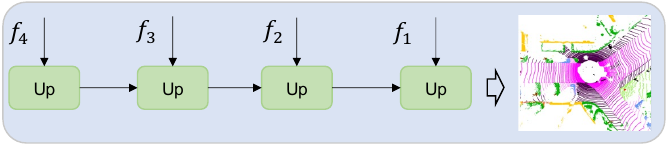}}
  \centerline{(b) Segmentation Head.}\medskip
\end{minipage}
\hfill
\begin{minipage}[b]{0.45\linewidth}
  \centering
  \centerline{\includegraphics[width=0.9\linewidth]{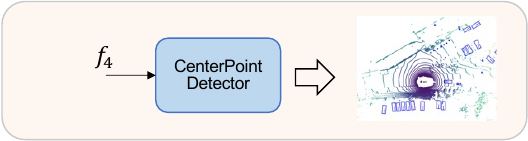}}
  \centerline{(c) CenterPoint Detector.}\medskip
\end{minipage}

\vspace{-0.5cm}
\caption{(a) Architecture of the LinK-based backbone; (b) the constructed network for 3D semantic segmentation; (c) the constructed network for 3D object detection.}
\label{fig:architecture}

\end{figure*}

\subsubsection{Block Based Aggregation}\label{sec:blockagg}

The above linear kernel enables reusing the overlapped area. Then a new question emerges: how to set up the overlap area? Motivated by ViT~\cite{dosovitskiy2020image}, we partition the entire input space into several non-overlapped blocks. Specifically, for the input scene $P\in \mathbb{Z}^{N\times 3}$, we quantize each voxel $p$'s coordinate with a block size $s$ and compute the corresponding hash code $l$ as follows:
\vspace{-0.1cm}
\begin{equation}
		l=Hash(\lfloor \frac{p(0)}{s} \rfloor, \lfloor \frac{p(1)}{s} \rfloor, \lfloor \frac{p(2)}{s} \rfloor).
\end{equation}
\noindent
Voxels owning the same hash code will be grouped into the same block. The block collection is denoted as $\mathbf{B}=\{B_0, B_1, ..., B_m\}$. Based on Formula~\ref{eq:overlap}, we conduct block-wise proxy aggregation for reusing as follows:
\vspace{-0.2cm}
\begin{equation}
	\begin{split}
		\left \{
			\begin{array}{ll}
			    f^{(0)}_{B_i} = \sum_{x\in B_{i}}{k^{(0)}(x)\cdot f_{x}},  \\
			    f^{(1)}_{B_i} = \sum_{x\in B_{i}}{k^{(1)}(x)\cdot f_{x}}.
			\end{array}
		\right.
	\end{split}
\end{equation}

\noindent
The $f^{(0)}_{B_i}$ and $f^{(1)}_{B_i}$ carry information with a receptive field of $s^3$. 

To expand the receptive field, we query the neighboring blocks' aggregation results for each block with a query range of $r^3$. For block $B_i$, denoting its neighboring block set as $\mathbb{B}_i$, then we compute the expanded block aggregation as 
\vspace{-0.2cm}
\begin{equation}
	\begin{split}
			\left \{
			\begin{array}{ll}
				f^{(0)}_{\mathbb{B}_i} = \sum_{j\in \mathbb{B}_{i}}{f^{(0)}_{B_j}},  \\
				f^{(1)}_{\mathbb{B}_i} = \sum_{j\in \mathbb{B}_{i}}{f^{(1)}_{B_j}}.	
			\end{array}
			\right.
	\end{split}
\end{equation}

\noindent
The $g^{(0)}_{\mathbb{B}_i}$ and $g^{(1)}_{\mathbb{B}_i}$ carry information with a receptive field of $(r\times s)^3$. For voxel $x$ within the block $B_i$, its final feature is updated by 

\begin{equation}
	g_x = \frac{1}{N_{\mathbb{B}_i}}[g^{(0)}_{\mathbb{B}_i}\cdot k^{(0)}(x)+g^{(1)}_{\mathbb{B}_i}\cdot k^{(1)}(x)]
\end{equation}

\noindent
$N_{\mathbb{B}_i}$ is the count of non-empty voxels. The voxel $x$ aggregates the information within a $(r\times s)^3$ area in a convolution-like manner.

\begin{figure}[t]
   \centering
   \includegraphics[width=0.8\linewidth]{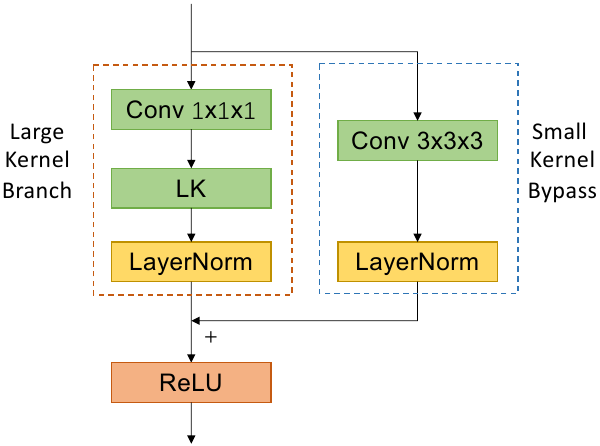}
\vspace{-0.3cm}
  \caption{Structures of LinK Module. The LK branch is responsible for a large kernel size while the Conv$3\times3\times3$ bypass makes up for the delicate local structures.}
  \label{fig:linkmodule}
  \vspace{-0.3cm}
\end{figure}

\subsubsection{Enhancements to the Kernel Generation}

To enhance the representation, two simple strategies from different views are proposed to improve kernel generation.

\noindent
{\bf Learnable Frequency for the Activation} For enhancing the model capability, we introduce two improvements to the cos-based and sin-based activation: channel-wise learnable parameters $\alpha$ to adjust the frequency, and an identity term $+x$ to preserve spatial information. The augmented activation function $\Psi'(*)$ is as follows:

\begin{equation}
    \Psi'(x)=\Psi(\alpha\cdot x)+x
\end{equation}

\noindent
{\bf Group Sharing Weight} The effective receptive field is significantly enlarged by block-based aggregation. Since each voxel only contributes once to the kernel generation, a large kernel range makes learning the kernel of each offset difficult. To facilitate the optimization, we adopt a group-sharing policy. Specifically, for the $C_{in}$ input channels, we only generate kernels of $\frac{C_{in}}{\#groups}$ channels and let every ${\#groups}$ channels share the same weight. Thus every weight would have more chances to be updated. We use $\#groups=2$ in practice.

\begin{table*}[t]
\footnotesize
\centering
    \caption{Results on the test phase of nuScenes Detection. {\bf Bold}: best results. * denotes using TTA.}
    \vspace{-0.3cm}
\setlength{\tabcolsep}{0.2cm}{
\label{tab:dettest}
\begin{tabular}{c|c|c|c|ccccccccccc}

\hline
 Methods        & Source& NDS  & mAP  &  \rotatebox{90}{car}  & \rotatebox{90}{truck} & \rotatebox{90}{bus}  & \rotatebox{90}{trailer} & \rotatebox{90}{\makecell[c]{construction\\\_vehicle}} & \rotatebox{90}{pedestrian} & \rotatebox{90}{motorcycle} & \rotatebox{90}{bicycle} & \rotatebox{90}{traffic\_cone} & \rotatebox{90}{barrier} \\
\hline
    PointPillars~\cite{pointpillars}   & {\itshape CVPR19}      & 45.3          & 30.5          & 68.4          & 23.0          & 28.2          & 23.4          & 4.1                   & 59.7          & 27.4          & 1.1           & 30.8          & 38.9          \\
    3DSSD~\cite{yang20203dssd}          & {\itshape CVPR20}      & 56.4          & 42.6          & 81.2          & 47.2          & 61.4          & 30.5          & 12.6                  & 70.2          & 36.0          & 8.6           & 31.1          & 47.9          \\
    CenterPoint~\cite{centerpoint}    & {\itshape CVPR21}      & 65.5          & 58.0          & 84.6          & 51.0          & 60.2          & 53.2          & 17.5                  & 83.4          & 53.7          & 28.7          & 76.7          & 70.9          \\
    HotSpotNet~\cite{chen2020object}     & {\itshape ECCV20}      & 66.0          & 59.3          & 83.1          & 50.9          & 56.4          & 53.3          & 23.0                  & 81.3          & 63.5          & 36.6          & 73.0          & 71.6          \\
    TransFusion-L~\cite{transfusion}    & {\itshape CVPR22}      & 70.2          & 65.5          & 86.2          & \textbf{56.7}          & 66.3          & 58.8          & 28.2                  & 86.1          & 68.3          & 44.2          & \textbf{82.0}          & \textbf{78.2}          \\
    Focals Conv~\cite{chen2022focal}    & {\itshape CVPR22}      & 70.0          & 63.8          & \textbf{86.7} & 56.3          & \textbf{67.7} & 59.5          & 23.8                  & \textbf{87.5} & 64.5          & 36.3          & 81.4          & 74.1          \\
    LargeKernel~\cite{lk3d}    & {\itshape arXiv22}     & 70.5          & 65.3          & 85.9          & 55.3          & 66.2          & 60.2          & 26.8                  & 85.6          & 72.5          & 46.6          & 80.0          & 74.3          \\
    \hline
    LinK    &            {\itshape Ours}    & \textbf{71.0} & \textbf{66.3} & 86.1          & 55.7          & 65.7          & \textbf{62.1} & \textbf{30.9}         & 85.8          & \textbf{73.5}          & \textbf{47.5} & 80.4          & 75.5          \\
    \hline\hline
    VISTA*~\cite{deng2022vista}         & {\itshape CVPR22}     & 70.4          & 63.7          & 84.7          & 54.2          & 64.0          & 55.0          & 29.1                  & 83.6          & 71.0          & 45.2          & 78.6          & 71.8          \\
    UVTR-LiDAR*~\cite{li2022unifying}    & {\itshape NeurIPS22} & 69.7          & 63.9          & 86.3          & 52.2          & 62.8          & 59.7          & 33.7                  & 84.5          & 68.8          & 41.1          & 74.7          & 74.9          \\
    MDRNet*~\cite{huang2022rethinking}      & {\itshape arXiv22}     & 72.8          & 68.4          & \textbf{87.9} & 58.5          & 67.3          & 64.1          & 30.2                  & \textbf{89.0} & 77.0          & 50.7          & \textbf{85.0} & 74.7          \\ 
    LargeKernel3D*~\cite{lk3d} &{\itshape arXiv22}     & 72.8          & 68.8          & 87.3          & 59.1          & 68.5          & 65.6          & 30.2                  & 88.3          & 77.8          & 53.5          & 82.4          & 75.0          \\
    \hline
    LinK*   & {\itshape  Ours}             & \textbf{73.4} & \textbf{69.8} & 87.3          & \textbf{60.2} & \textbf{69.8} & \textbf{65.9} & \textbf{34.0}         & 88.2          & \textbf{78.8} & \textbf{54.3} & 83.0          & \textbf{76.8} \\
    \hline\hline

\end{tabular}
}

\end{table*}

\subsection{Network Structure} 
\label{sec:task}
\subsubsection{LinK Module}

Since LinK aggregates spatial information in a depth-wise manner, we apply a $1^3$ convolution to the input feature before sending it to the LinK operator to introduce channel mixing~\cite{chollet2017xception,tolstikhin2021mlp}. Meanwhile, following the previous practice~\cite{replknet,lk3d}, we append a parallel $3^3$ convolutional branch to preserve detailed structures. This operation also stabilizes the optimization process. The resulting architecture is illustrated in Fig~\ref{fig:linkmodule}. Unlike the choices in LargeKernel3D~\cite{lk3d}, we do not adopt a dilation $>1$ for the $3^3$ branch. And we replace the BatchNormalization~\cite{ioffe2015batch} with LayerNormalization~\cite{ba2016layer} to reinforce those informative channels.

\subsubsection{Applications in Perception Tasks}

LinK is incorporated into two essential perception tasks: 3D object detection and 3D semantic segmentation. We choose two representative architectures for the two tasks and directly replace their SparseConv-based backbone with the LinK-based backbone and keep the original design of their segmentation head and detector. Detailed architectures are shown in Fig~\ref{fig:architecture}.

\section{Experiments}\label{sec:experiments}
To verify the effectiveness of our method and explore its characteristics, we conduct extensive experiments in this section. The whole project is implemented upon three software architectures: PyTorch~\cite{paszke2019pytorch}, TorchSparse~\cite{tang2022torchsparse}, and SpConv~\cite{spconv}. All experiments are conducted on a server with 4 RTX 3090 GPUs. We present two most representative datasets in this section. Results on more datasets can be found in the supplementary material.

 \begin{figure}[tb]
   \centering
\includegraphics[width=0.8\linewidth]{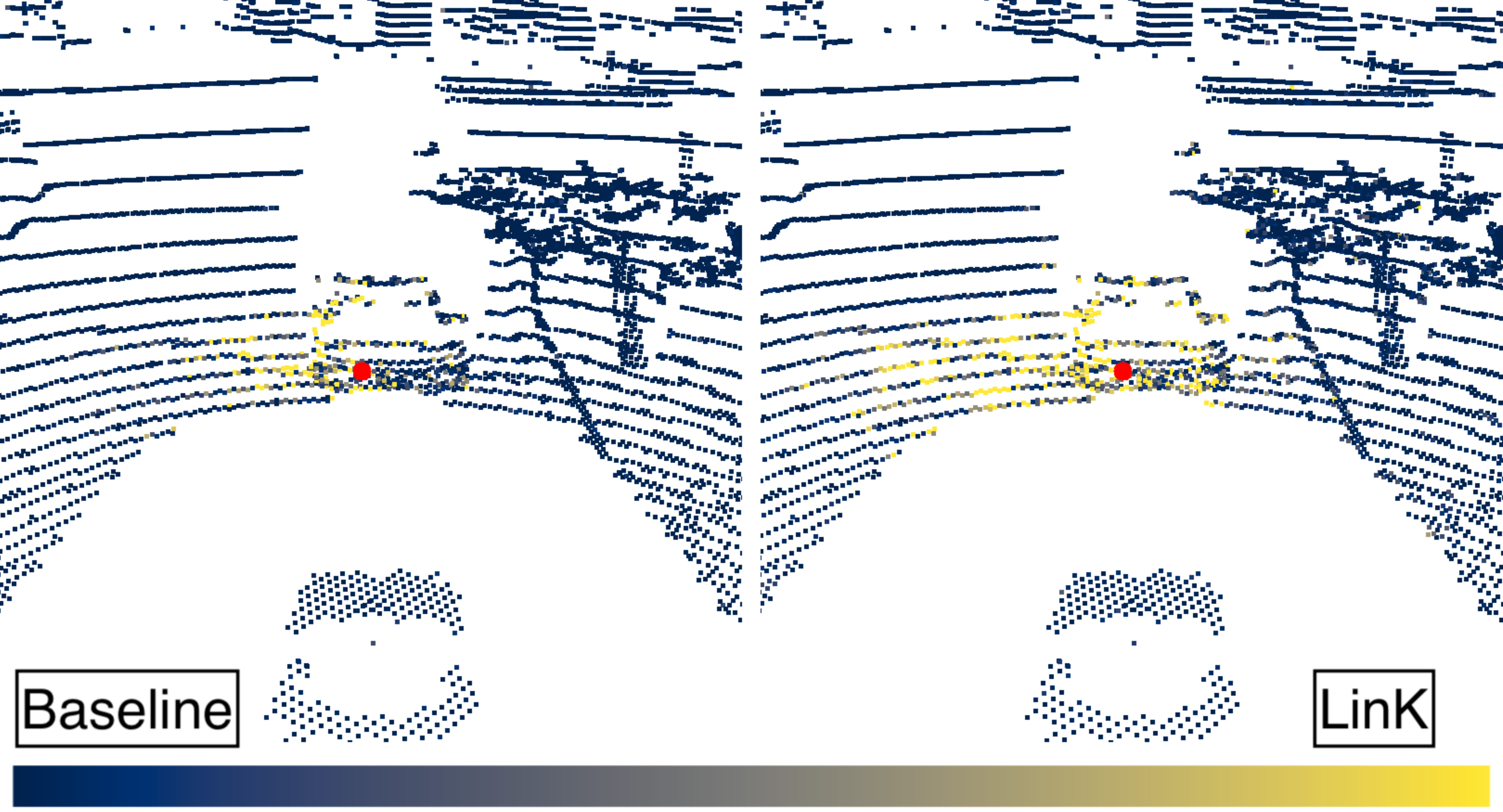}
 \caption{The effective receptive field (ERF) of the detection. The brightness indicates the degree of activation. LinK enjoys a wider-range perception.}
 \vspace{-0.2cm}
 \label{fig:deterf}
 \end{figure}

\begin{table*}[t]
\footnotesize
\centering
\caption{SemanticKITTI test results. \textcolor[rgb]{1,0,0}{Red}: surpassing the baseline; {\bf bold}: best results; 'P': point cloud; 'R': range map; 'V': voxel.}
	\vspace{-0.3cm}
\setlength{\tabcolsep}{0.06cm}{
\label{tab:segtest}
\begin{tabular}{c|c|c|ccccccccccccccccccc}
    \hline
Method    &Input&mIoU    & \rotatebox{90}{Car}  & \rotatebox{90}{Bicycle} & \rotatebox{90}{Motorcycle} & \rotatebox{90}{Truck} & \rotatebox{90}{Other-vehicle} & \rotatebox{90}{Person} & \rotatebox{90}{Bicyclist} & \rotatebox{90}{Motorcyclist} & \rotatebox{90}{Road} & \rotatebox{90}{Parking} & \rotatebox{90}{Sidewalk} & \rotatebox{90}{Other-ground} & \rotatebox{90}{Building} & \rotatebox{90}{Fence} & \rotatebox{90}{Vegetation} & \rotatebox{90}{Trunk} & \rotatebox{90}{Terrain} & \rotatebox{90}{Pole} & \rotatebox{90}{Traffic-sign} \\
    \hline
    RandLA-Net~\cite{randla}   & P     & 53.9          & 94.2                                   & 26.0                                       & 25.8                                          & 40.1                                     & 38.9                                             & 49.2                                      & 48.2                                         & 7.2                                             & 90.7                                    & 60.3                                       & 73.7                                        & 20.4                                            & 86.9                                        & 56.3                                     & 81.4                                          & 61.3                                     & 66.8                                       & 49.2                                    & 47.7                                            \\
    RangeNet++~\cite{milioto2019rangenet++}   & R     & 52.2          & 91.4                                   & 25.7                                       & 34.4                                          & 25.7                                     & 23.0                                             & 38.3                                      & 38.8                                         & 4.8                                             & 91.8                                    & 65.0                                       & 75.2                                        & 27.8                                            & 87.4                                        & 58.6                                     & 80.5                                          & 55.1                                     & 64.6                                       & 47.9                                    & 55.9                                            \\
    SqueezeSegV3~\cite{xu2020squeezesegv3} & R     & 55.9          & 92.5                                   & 38.7                                       & 36.5                                          & 29.6                                     & 33.0                                             & 45.6                                      & 46.2                                         & 20.1                                            & 91.7                                    & 63.4                                       & 74.8                                        & 26.4                                            & 89.0                                        & 59.4                                     & 82.0                                          & 58.7                                     & 65.4                                       & 49.6                                    & 58.9                                            \\
    SalsaNext~\cite{cortinhal2020salsanext}    & R     & 59.5          & 91.9                                   & 48.3                                       & 38.6                                          & 38.9                                     & 31.9                                             & 60.2                                      & 59.0                                         & 19.4                                            & 91.7                                    & 63.7                                       & 75.8                                        & 29.1                                            & 90.2                                        & 64.2                                     & 81.8                                          & 63.6                                     & 66.5                                       & 54.3                                    & 62.1                                            \\
    SPVNAS~\cite{spvnas}       & P+V   & 67.0          & 97.2                                   & 50.6                                       & 50.4                                          & 56.6                                     & 58.0                                             & 67.4                                      & 67.1                                         & 50.3                                            & 90.2                                    & 67.6                                       & 75.4                                        & 21.8                                            & 91.6                                        & 66.9                                     & 86.1                                          & 73.4                                     & 71.0                                       & 64.3                                    & 67.3                                            \\
    Cylinder3D~\cite{cylinder3d}     & V     & 67.8          & 97.1                                   & 67.6                                       & 64.0                                          & 59.0                                     & 58.6                                             & 73.9                                      & 67.9                                         & 36.0                                            & 91.4                                    & 65.1                                       & 75.5                                        & 32.3                                            & 91.0                                        & 66.5                                     & 85.4                                          & 71.8                                     & 68.5                                       & 62.6                                    & 65.6                                            \\
    (AF)2-S3Net~\cite{cheng20212}  & V     & 69.7          & 94.5                                   & 65.4                                       & \textbf{86.8}                                 & 39.2                                     & 41.1                                             & \textbf{80.7}                             & \textbf{80.4}                                         & \textbf{74.3}                                   & 91.3                                    & 68.8                                       & 72.5                                        & \textbf{53.5}                                   & 87.9                                        & 63.2                                     & 70.2                                          & 68.5                                     & 53.7                                       & 61.5                                    & \textbf{71.0}                                            \\
    DRINet~\cite{drinet}       & P+V   & 67.5          & 96.9                                   & 57.0                                       & 56.0                                          & 43.3                                     & 54.5                                             & 69.4                                      & 75.1                                         & 58.9                                            & 90.7                                    & 65.0                                       & 75.2                                        & 26.2                                            & 91.5                                        & 67.3                                     & 85.2                                          & 72.6                                     & 68.8                                       & 63.5                                    & 66.0                                            \\
    RPVNet~\cite{rpvnet}       & R+P+V & 70.3          & \textbf{97.6}                          & \textbf{68.4}                              & 68.7                                          & 44.2                                     & 61.1                                             & 75.9                                      & 74.4                                         & 73.4                                            & \textbf{93.4}                           & \textbf{70.3}                                       & \textbf{80.7}                               & 33.3                                            & \textbf{93.5}                               & \textbf{72.1}                            & \textbf{86.5}                                          & \textbf{75.1}                            & \textbf{71.7}                                       & \textbf{64.8}                                    & 61.4                                            \\
    \hline
    Mink(baseline)~\cite{choy20194d} &V& 68.0  & 97.1 & 51.8 & 56.4 & 43.3 & 56.8 & 70.2 & 75.7 & 51.8 & 89.9 & 67.8 & 74.8 & 32.9 & 91.5 & 66.5 & 86.2 & 74.6 & 71.0 & 63.5 & 70.0 \\
    	LinK(Ours)         & V     & \textbf{70.7} & \textcolor[rgb]{1,0,0}{97.4}                                   & \textcolor[rgb]{1,0,0}{58.4}                                       & \textcolor[rgb]{1,0,0}{56.6}                                          & \textcolor[rgb]{1,0,0}{52.9}                                     & \textbf{64.2}                                    & \textcolor[rgb]{1,0,0}{72.3}                                      & \textcolor[rgb]{1,0,0}{77.0}                                         & \textcolor[rgb]{1,0,0}{69.1}                                            & \textcolor[rgb]{1,0,0}{90.6}                                    & \textcolor[rgb]{1,0,0}{68.2}                                       & \textcolor[rgb]{1,0,0}{76.2}                                        & \textcolor[rgb]{1,0,0}{34.5}                                            & \textcolor[rgb]{1,0,0}{92.0}                                        & \textcolor[rgb]{1,0,0}{68.8}                                     & 85.7                                          & 74.3                                     & 70.5                                       & \textbf{64.8}                                    & 69.5                     \\                      
    \hline
\end{tabular}}

\end{table*}

\subsection{3D Object Detection}

\subsubsection{Dataset}

We evaluate the 3D detection performance on the widely used benchmark, nuScenes~\cite{nuscenes}, a public dataset for autonomous driving. It is collected from challenging urban scenes, consisting of 1000 annotated sequences. Among them, 700 sequences are used as the training phase, 150 as validation, and the remaining 150 as the test phase for online evaluation. The bounding boxes are labeled with not only categories but also some attributes like the velocity, scale, orientation and translation. Besides the point cloud data from LiDAR, it also provides the $360^{\circ}$ image modality from cameras and signal from the Radar sensor. In this paper, we only utilize the point cloud from LiDAR. The evaluation metrics include the mAP and a dataset-related NDS, i.e., nuScenes Detection Score. The NDS measures the comprehensive performance by combining the mAP and other attributes in a weighted manner.

\subsubsection{Implementation Details}

The kernel configuration for the detection is $\{r=3, s=7\}$. We implement the detection codebase by replacing the original backbone of CenterPoint~\cite{centerpoint} with a LinK-based backbone. For a fair comparison,  we keep all the original hyperparameters in CenterPoint~\cite{centerpoint} to train our network. Following the common practice, the training phase is augmented with the CBGS~\cite{cbgs} and gt-sampling~\cite{yan2018second} strategy to balance the long-tail issues. During test and validation, we follow LargeKernel3D~\cite{lk3d} to report the plain inference result and the test-time augmented (TTA, including flip and rotation) inference results.

\subsubsection{Results}
\vspace{-0.1cm}
We compare the detection results with many representative methods. All the test results in Table~\ref{tab:dettest} are obtained through public sources, like published papers or competition websites. Until the time of paper submission, our method ranks \textcolor[rgb]{1,0,0}{1st} on the public LiDAR detection leaderboard. Especially for the NDS metric, which has been stuck in 72.7\%$\sim$ 72.8\% for more than one year, we are the first method to obtain a result higher than 73\% on the LiDAR track, which confirms the superiority of our large kernel method. Table~\ref{tab:detval} shows the comparisons on the validation phase. We achieve consistent improvement in both settings. The effective receptive field is shown in Fig~\ref{fig:deterf}, and our approach enjoys a larger ERF than the baseline method. More qualitative results are available in the supplementary material.

\begin{table}[t]
\footnotesize
\begin{minipage}[c]{0.21\textwidth}

\centering
\setlength{\tabcolsep}{0.1cm}{

	\captionof{table}{ Val@nuScenes Det.}
	\vspace{-0.3cm}
	\label{tab:detval}  
\begin{tabular}{c|cc}

\hline
 Methods      & NDS  & mAP  \\
\hline
    CBGS~\cite{cbgs}       &   62.6       &   51.4        \\
%    3DSSD            &     & \\
    CenterPoint~\cite{centerpoint}     &  66.4         & 59.0 \\
    HotSpotNet~\cite{chen2020object}     &       66.0    & 59.5 \\
    TransFusion-L~\cite{transfusion}  & 66.8  & 60.0  \\
    Focals Conv~\cite{chen2022focal}     &    68.1      & 61.2 \\
    LargeKernel3D~\cite{lk3d}    &     69.1      & 63.3       \\
    \hline
    LinK & {\bf69.5} & {\bf63.6}  \\
    \hline
\end{tabular}}

\hspace{1cm}
\end{minipage}
\begin{minipage}[c]{0.34\textwidth}
\centering

\setlength{\tabcolsep}{0.1cm}{   
	\captionof{table}{Val@SemKITTI Seg.}
	\vspace{-0.3cm}
\label{tab:segval}
\begin{tabular}{c|c}

\hline
 Methods      & mIoU    \\
\hline
    RandLA-Net~\cite{randla}       & 57.1               \\
    RangeNet++~\cite{milioto2019rangenet++}             &  57.3   \\
    SPVNAS~\cite{spvnas}     &   64.7          \\
    Cylinder3D~\cite{cylinder3d}     &  63.8           \\
    RPVNet~\cite{rpvnet}  &  65.5    \\
    Mink~\cite{choy20194d}  &  66.1    \\
    \hline
    LinK & \textbf{67.5}   \\
    \hline
\end{tabular}
}

\end{minipage}

\end{table}

\subsection{3D Semantic Segmentation}
\subsubsection{Dataset}

We evaluate the semantic segmentation performance on the SemanticKITTI~\cite{semantickitti}. It is a large-scale (including more than 43000 scans, and each scan has more than 100k points) autonomous dataset constructed by labeling the odometry dataset KITTI~\cite{kitti} with 20 categories of semantic masks. The dataset contains 22 sequences and has been officially partitioned into three phases: sequences [00-07, 09, 10] as the training phase, sequence 08 as the validation phase, and the rest sequences [11-21] as the online testing benchmark. The evaluation metric is the mean Intersection over Union.

\subsubsection{Implementation Details}
The kernel configuration for the segmentation is $\{r=2, s=3\}$. We train the network for 25 epochs in total. Following SPVNAS~\cite{spvnas}, we adopt an initial learning rate of 2.4e-1 and adjust it using the cosine scheduler. The optimizer for updating learnable parameters is SGD. All experiments are run with 4 GPUs in parallel, and the batch size in each GPU is 2. To deal with different object scales, we follow previous works~\cite{cylinder3d,rpvnet} to introduce the Lovasz loss~\cite{berman2018lovasz} to cooperate with the original cross-entropy loss. The size for voxelization is 0.05m. We preserve 80,000 points for each scan to train the network. During the validation and test processes, we report the result of direct inference and the TTA results. Due to the severe long-tail problem, this dataset is very sensitive to some small categories, like the bicycle, person, and motorcyclist. Some model-ensemble techniques and instance-level augmentations (like conducting the Copy-Paste and Cut-Mix on some small categories) contribute significantly to the final results. Because there are no comprehensive works to conclude these tricks in a unified manner, we do not adopt these augmentations to prevent from shadowing the essence of model design.

\subsubsection{Results}
The LinK for segmentation is implemented by only replacing the Mink's~\cite{choy20194d} encoder with a LinK-based encoder. So, the Mink serves as a direct baseline. Table~\ref{tab:segtest} shows the test results on SemanticKITTI~\cite{semantickitti}. Most baseline results are obtained from the public sources. For Mink~\cite{choy20194d}, we reproduce it with the same configurations as ours and surprisingly find that this basic architecture hits performance on par with other complex architectures. When combined with the LinK module to enlarge the receptive field, we achieve an improvement of 2.7\% in mIoU. Table~\ref{tab:segval} shows the results of the validation phase. Visulaizations are provided in the supplementary material.

\begin{table}[t]
\footnotesize
    \centering
            \caption{Performance on different scale objects.}
            \label{tab:differentsize}
            	\vspace{-0.3cm}
    \setlength{\tabcolsep}{0.15cm}{
    \begin{tabular}{c|c|c|c|c|c}
         \hline
         \multirow{2}{*}{Category}& \multirow{2}{*}{Size($m^3$)} & \multicolumn{2}{c|}{Detection}& \multicolumn{2}{c}{Segmentation}  \\ \cline{3-6}
         &&\makecell[c]{Center\\Point}&+LinK&Mink &+LinK \\ \hline
         Truck&$6\times2\times2$&51.0&\textcolor[rgb]{0,0,1}{(+4.7)}55.7&43.3&\textcolor[rgb]{0,0,1}{(+9.6)}52.9 \\ \hline
         Person&$0.4\times0.4\times2$&83.4&\textcolor[rgb]{0,0,1}{(+2.4)}85.8&70.2&\textcolor[rgb]{0,0,1}{(+2.1)}72.3 \\ \hline
    \end{tabular}}
    \vspace{-0.2cm}
\end{table}

\subsection{Overhead Analysis}
We measure the number of parameters and the inference speed to evaluate LinK's practicability in Table~\ref{tab:segoverhead} and Table~\ref{tab:detoverhead}. As shown in Table~\ref{tab:segoverhead}, we achieve a better mIoU by consuming fewer computation resources than the naive Conv$7\times7\times7$ operator in the segmentation task. We measure the performance on a single 3090 GPU with bs=1.

 \begin{table}
 \footnotesize
    \centering
        \caption{Time and parameters analysis for semantic segmentation.}
        	\vspace{-0.3cm}
    \setlength{\tabcolsep}{0.15cm}{
	\label{tab:segoverhead}
    \begin{tabular}{c|cc|c}
         \hline
         Methods &$$Parameters&Runtime(ms)&mIoU(\%) \\ \hline
         Mink&8.5M &69&66.1\\
	Conv$7\times7\times7$ &21.75M&139&66.8 \\ \hline
         LinK&10.75M&87&67.5 \\ \hline
    \end{tabular}}
    \vspace{-0.2cm}
\end{table}

\begin{table}[t]
 \footnotesize
    \centering
        \caption{Time and parameters analysis for 3D object detection.}
        \vspace{-0.3cm}
    \setlength{\tabcolsep}{0.15cm}{
	\label{tab:detoverhead}
    \begin{tabular}{c|cc|cc}
         \hline
         Methods &$$Parameters&Runtime(ms)&mAP&NDS \\ \hline
         CenterPoint&8.6M &73&59.0&66.4\\ \hline
         LinK&10.3M&109&60.3&67.7 \\ \hline
    \end{tabular}}
    \vspace{-0.3cm}
\end{table}

\subsection{Ablation Studies}
\noindent
{\bf How large kernel play a role in 3D perception?} We investigate this question through two ways. First, according to Fig~\ref{fig:deterf}, LinK-based backbone produces a wider range of ERF. Second, as shown in Table~\ref{tab:differentsize}, big objects benefit more from the large kernel. In conclusion, we think the large kernel enhances the network's ability to model different scales objects more effectively with fewer layers.

\vspace{0.1cm}

\noindent
{\bf The branch combinations in LinK module.} We ablate the construction of the LinK-based architecture from two aspects: (1). does the large kernel really contribute positively? (2). its comparisons with the standard large kernel (Conv$7\times7\times7$). According to Table~\ref{tab:branchs}, both the standard kernel of $7\times7\times7$ and the LinK improve the baseline with the aid of a $3\times3\times3$ bypass branch, which verifies the effectiveness of a large kernel. Meanwhile, the LinK outperforms the standard large kernel with a non-trivial margin. When removing the ResBlock branch, i.e., the backbone only consists of large kernels, the network still hits a high mAcc. This implies the large kernel's ability in modeling large-scale objects.
\vspace{0.1cm}

\begin{figure}[t]
   \centering
   \includegraphics[width=0.7\linewidth]{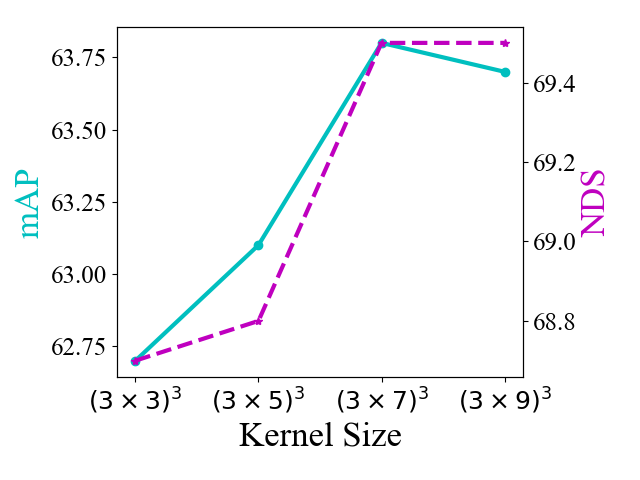}
\vspace{-0.5cm}
  \caption{Detection performance with different kernel sizes.}
  \label{fig:perfwithsize}
\end{figure}

\begin{table}
\footnotesize
    \centering
            \caption{Ablations on the backbone structures. 'RB': ResBlock,  'LKB': large kernel branch, 'LK': how to implement large kernel.}
            	\vspace{-0.3cm}
    \setlength{\tabcolsep}{0.15cm}{
    \begin{tabular}{c|c|c|c|c}
         \hline
         \multirow{2}{*}{RB} & \multicolumn{2}{c|}{LKB}  & \multirow{2}{*}{mIoU(\%)$\uparrow$}& \multirow{2}{*}{mAcc(\%)$\uparrow$} \\ \cline{2-3}
         &Bypass&LK& &\\

         \hline
        \multirow{7}{*}{ \checkmark}&&&66.1&72.4 \\ \cline{2-5}
         &\checkmark&&66.4& 72.9\\ \cline{2-5}
          & \checkmark &Conv$7\times7\times7$&66.8& 72.3\\ \cline{2-5}
         &&Conv$7\times7\times7$ &66.1 &72.3\\ \cline{2-5}
         &&LinK& 65.6& 72.3\\ \cline{2-5}
         &\checkmark&LinK& {\bf67.5} &{\bf74.7}\\ \hline
         &\checkmark&LinK& 65.7 &73.6\\ \hline
    \end{tabular}}
    \label{tab:branchs}
\end{table}

\noindent
{\bf The kernel size.} We explore the influence of LinK's kernel size in the detection task. As shown in Fig~\ref{fig:perfwithsize}, the performances increase along with the kernel size within a range and saturate when the kernel size is larger than $(3\times7)^3$.

\vspace{0.1cm}

\begin{table}[t]
\footnotesize
    \centering
        \caption{Ablations on the two augmentations of kernel.}
        \vspace{-0.3cm}
    \setlength{\tabcolsep}{0.1cm}{
    \begin{tabular}{c|cc|c}
         \hline
	$r\times s$&\makecell[c]{Learnable\\ Frequency}&          \makecell[c]{Group\\ Sharing} &mIoU(\%) \\ \hline
	\multirow{ 3}{*}{$2\times 3$}&&& 67.3\\ \cline{2-4}
          &\checkmark&&\textbf{67.5}\\ \cline{2-4}
	&&\checkmark&67.1\\ \hline
	\hline
	
	\multirow{ 3}{*}{$3\times 5$}&&& 66.2\\ \cline{2-4}
          &\checkmark&&66.8\\ \cline{2-4}
	&&\checkmark&{\bf67.5}\\ \hline
                                       
    \end{tabular}}
    \label{tab:cyclegroup}
    
    \vspace{-0.3cm}
\end{table}

\noindent
{\bf The two strategies to augment kernel.} We validate the effectiveness of the two augmentations to the kernel generation. Table~\ref{tab:cyclegroup} shows that introducing the learnable frequency works well with a relatively small kernel, and the grouping sharing weight performs better with a larger kernel.

\section{Conclusion}

\label{sec:conclusion}
A large receptive field is essential in computer vision tasks. In this paper, we have posed a linear kernel generator, LinK, to enlarge the effective receptive field for 3D perception tasks at the cost of moderate computations. Extensive experimental results on the detection and segmentation demonstrate the effectiveness of the proposed LinK, and we achieve consistent improvements over baselines. In the future, we will work on generalizing this method to more basic models like the Transformer and dynamic convolution.

\noindent
{\bf Acknowledgements} This work is supported by the National Key R$\&$D Program of China (No. 2022ZD0160900), the National Natural Science Foundation of China (No. 62076119, No. 61921006, No. 62072232), the Fundamental Research Funds for the Central Universities (No. 020214380091), and the Collaborative Innovation Center of Novel Software Technology and Industrialization.

%%%%%%%%% REFERENCES

% supplementary
% \newpage

\appendix

%%%%%%%%% BODY TEXT - ENTER YOUR RESPONSE BELOW

\section{More implementation details and results}
\subsection{Detection}
\noindent
{\bf Training Process} Following common practice~\cite{cbgs,lk3d}, the reported validation results are obtained through training on the train split, and the results on test set are obtaining through training on the train+val split. The subset for training (train or train+val) are augmented using the CBGS strategy, which balances the sample distribution. Meanwhile, a gt-sampling strategy~\cite{yan2018second} is adopted to enhance object-level balance during training. Our network are trained with CBGS+gt-sampling for 15 epochs, and then finetuned by removing the gt-sampling for extra 5 epochs. Experiences in previous work indicate that such training policy benefits from the augmented dataset most while avoids overfitting the synthetic distribution. 

\noindent
{\bf Results} The TTA process for nuScenes contains the flipping and rotation. For flipping, we apply 4 operations: [no flip, x-axis, y-axis, x-axis+y-axis]. For the rotation, we adopt 7 angles, i.e., [$0^\circ$, $\pm6.25^\circ$, $\pm12.5^\circ$, $\pm25^\circ$]. Thus there are total 28 variants for each sample during inference. All the results for the same sample are reduced by a NMS process.

\subsection{Segmentation}
\noindent
{\bf Data Augmentation} The input for semantic segmentation is a 4-dimension tensor, consisting of the normalized coordinate of each point and the corresponding LiDAR reflection intensity. The coordinates are augmented with random flip along x-aixs or y-axis, random scaling within [0.95, 1.05], and random rotation within [0, $2\pi$). For the TTA process during inference, we apply the random augmentation for 12 times and average the results.

\section{Detailed layer architecture}

\begin{figure}[t]
  \centering
  \includegraphics[width=0.8\linewidth]{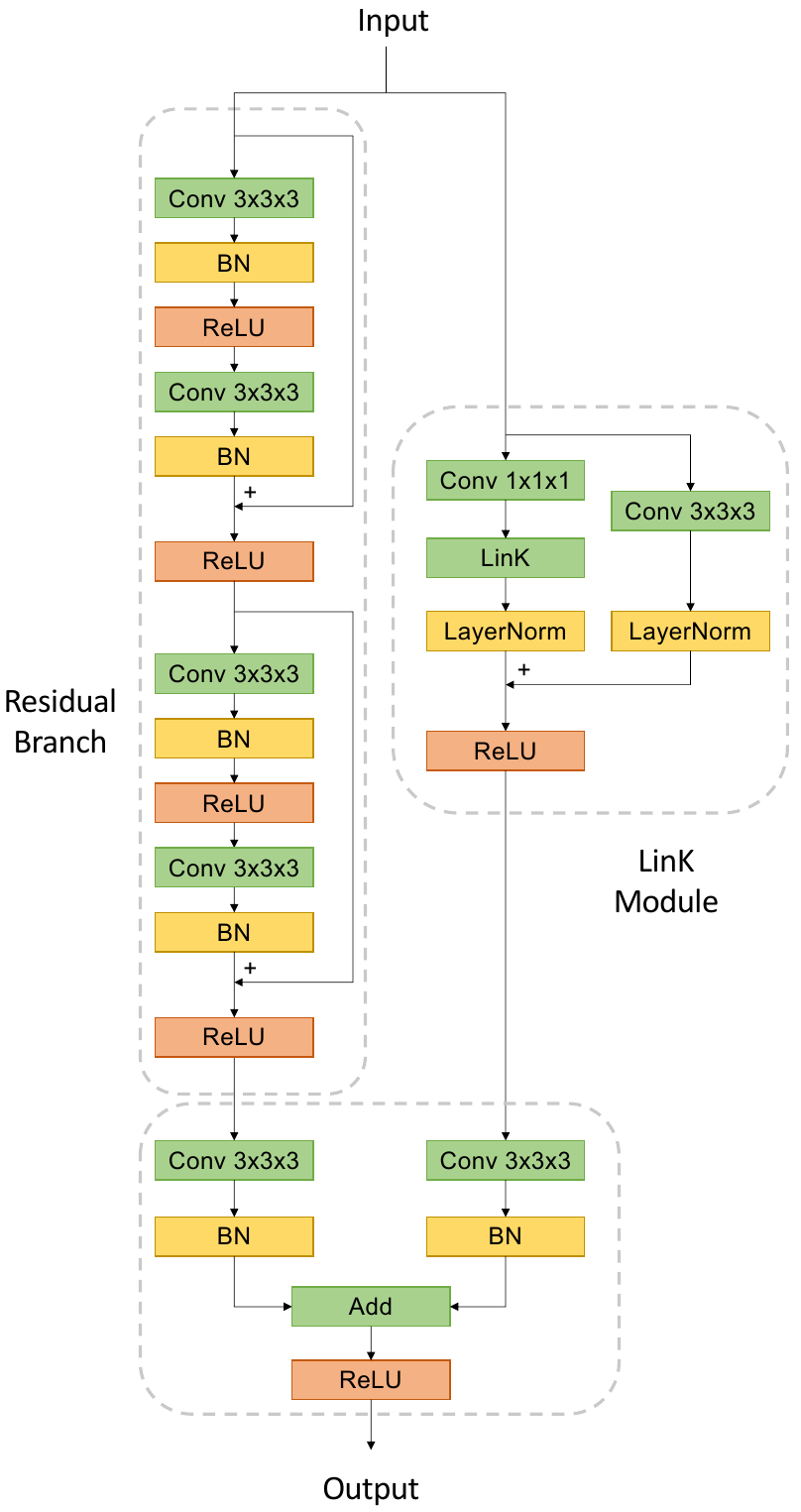}
  \caption{The detailed architecture of the encoder layer.}
  \label{fig:layer}
\end{figure}

Both of the segmentation and detection task share the same encoder design. The encoder starts with a Stem Block (Conv$3\times3\times3$+BN+ReLU+Conv$3\times3\times3$+BN+ReLU) and then appends with 4 downsample+parallel layers (Residual Branch $\|$ LinK Module). Each Residual Branch consists of two residual blocks. The detailed architecture of each parallel layer is depicted in Fig~\ref{fig:layer}. For segmentation task, the hidden dimensions for all the encoder layers are 64. For the detection, the hidden dimension is $[16,32,64,128]$.

\section{Results on more datasets}
To further explore the potential of LinK, we conduct experiments on other three benchmarks. For the 3D object detection, we first train the CenterPoint and LinK for 6 epochs on Waymo Detection~\cite{waymo}, the results on validation split are shown in Table~\ref{tab:waymodet}. We also train KITTI Detection~\cite{kitti} using CentePoint-KITTI~\cite{cp_kitti} and LinK under the same settings. The results is provided in Table~\ref{tab:kittidet}. The detection results on the two datasets further demonstrate the effectiveness of our large kernel design. 

For semantic segmentation, we design one more segmentation experiment on nuScenes~\cite{nuscenes}. According to Table~\ref{tab:nuscseg}, LinK achieves consistent improvement over the baseline.

\begin{table}[t]
\footnotesize
    \centering
        \caption{Different kernel sizes for segmentation. Without TTA.}
        \vspace{-0.3cm}
    \setlength{\tabcolsep}{0.1cm}{
    \begin{tabular}{c|c}
         \hline
	$r\times s$&\makecell[c]{mIoU(\%)@SemKITTI val} \\ \hline 
	$3\times 2$& 66.9 \\ 		
	$3\times 3$& 67.3 \\ 		
	$3\times 5$& 67.5 \\ 
	$3\times 7$& 67.2 \\ 
           \hline
                                       
    \end{tabular}}
    \label{tab:segsize}
    
    \vspace{-0.3cm}
\end{table}

\begin{table}[t]
\footnotesize
    \centering
        \caption{Validation on Waymo Detection. Trained for 6 epochs.}
        \vspace{-0.3cm}
    \setlength{\tabcolsep}{0.1cm}{
    \begin{tabular}{c|ccc|c}
         \hline
	$Methods$&Vehicle&Pedestrian&Cyclist&mAPH(\%)\\ \hline 
	$CenterPoint$& 63.4 & 59.5 &66.4& 63.4 \\ 	\hline	
	$\makecell[c]{+LinK}$& \textcolor[rgb]{0,0,1}{(+1.6)}65.0 & \textcolor[rgb]{0,0,1}{(+0.9)}60.4& \textcolor[rgb]{0,0,1}{(+2.0)}68.4& \textcolor[rgb]{0,0,1}{(+1.2)}64.6\\ 		
           \hline
                                       
    \end{tabular}}
    \label{tab:waymodet}
    
\end{table}

\begin{table}[t]
\footnotesize
    \centering
        \caption{Validation on KITTI Detection. (mAP)}
        \vspace{-0.3cm}
    \setlength{\tabcolsep}{0.1cm}{
    \begin{tabular}{c|ccc}
         \hline
	$Methods$&Easy&Moderate&Hard\\ \hline 
	$CenterPoint$& 71.2 & 61.7 & 58.1 \\ 	\hline	
	$\makecell[c]{+LinK}$& \textcolor[rgb]{0,0,1}{(+1.6)}72.8& \textcolor[rgb]{0,0,1}{(+2.1)}63.8& \textcolor[rgb]{0,0,1}{(+2.1)}60.2\\ 		
           \hline
                                       
    \end{tabular}}
    \label{tab:kittidet}
    
\end{table}

\begin{table}[t]
\footnotesize
    \centering
        \caption{Validation on nuScenes Segmentation. (\#channel=64, \#epoch=80, bs=16, voxel size=10cm)}
        \vspace{-0.3cm}
    \setlength{\tabcolsep}{0.1cm}{
    \begin{tabular}{c|ccc}
         \hline
	$Methods$&mIoU(\%)&mAcc(\%)&oAcc(\%)\\ \hline 
	$Mink$&74.8 &82.3 & 93.3\\ 	\hline	
	$+LinK$& \textcolor[rgb]{0,0,1}{(+1.6)}76.4& \textcolor[rgb]{0,0,1}{(+0.8)}83.1& \textcolor[rgb]{0,0,1}{(+0.4)}93.7 \\
           \hline
                                       
    \end{tabular}}
    \label{tab:nuscseg}
    
\end{table}

\section{More ablations}

{\bf Different kernel sizes in semantic segmentation task.} We report the segmentation results of more kernel sizes in Table~\ref{tab:segsize} and the performance saturates at $3\times5$.

{\bf Mirco-designs in LinK module.}  The default dilation in the bypass branch is 1, and we enlarge it to 2 in Table~\ref{tab:link_dilate_norm}. The comparison implies that there is no need to enlarge the receptive field of bypass branch. Furthermore, the norm type in LinK does not have an impact on the segmentation results.

\section{Visualizations}

\subsection{Kernel weight Distributions}
\begin{figure}[tb]
\begin{minipage}[b]{.9\linewidth}
   \centering
   \centerline{\includegraphics[width=0.9\linewidth]{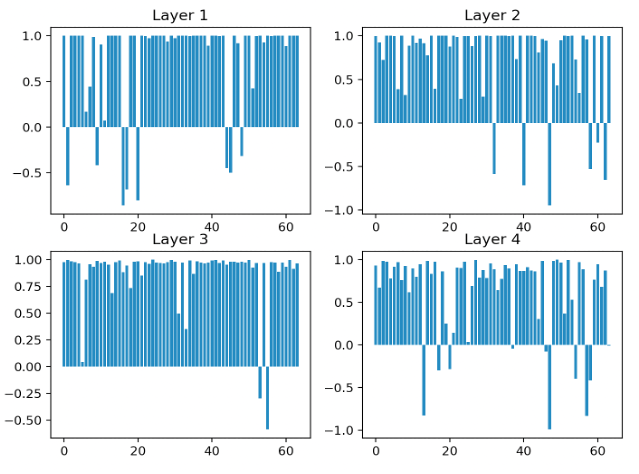}}
   
   \centerline{(a) w/o Group Sharing}\medskip
 \end{minipage}
 \hfill
 \begin{minipage}[b]{0.9\linewidth}
   \centering
     \centerline{\includegraphics[width=0.86\linewidth]{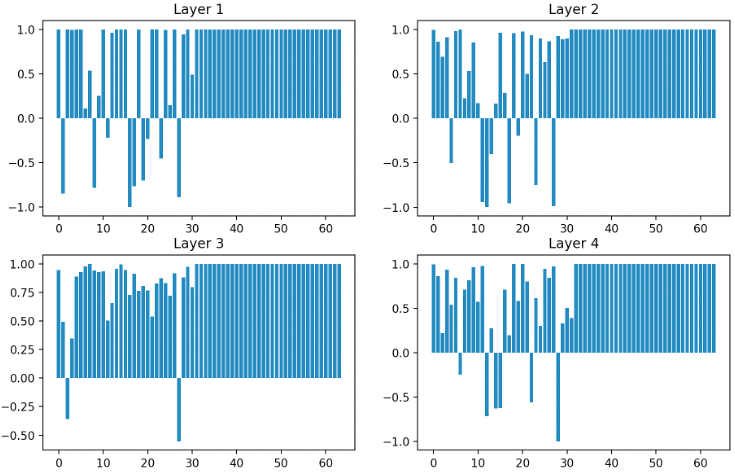}}

  \centerline{(b) Group Sharing}\medskip
 \end{minipage}
 \vspace{-0.5cm}
 \caption{Activations in different channels. For (b), the first 32 channels are repeated to serve as 64-channel weights. The latter 32 channels for the Group Sharing does not participate in any forward or backward process.}
 \label{fig:channel}
 \end{figure}

\noindent
{\bf Channel Distribution}
To explore the effect of the group sharing strategy, we analyze the weight distribution in channel dimension. According to Fig~\ref{fig:channel}, before adopting the Group Sharing strategy, the low-level channels are optimized insufficiently, since the variations among channels are very small. After adopting the Group Sharing policy, the low-level channel are activated significantly, which verifies the effectiveness of Group Sharing.

\begin{figure}[tb]
\begin{minipage}[b]{.49\linewidth}
   \centering
   \centerline{\includegraphics[width=0.9\linewidth]{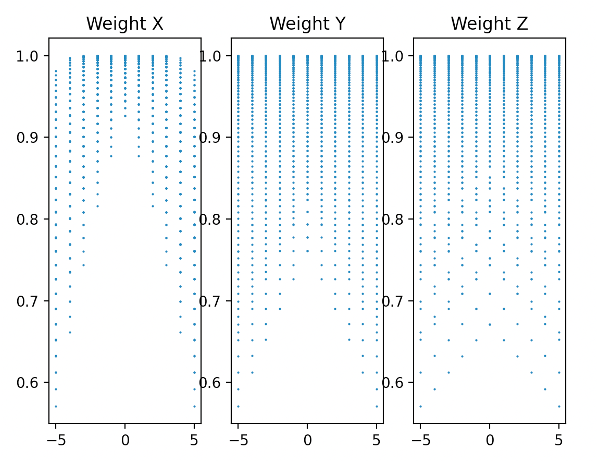}}
   
   \centerline{(a) $\Psi(x)=\cos(x)$}\medskip
 \end{minipage}
 \hfill
 \begin{minipage}[b]{0.49\linewidth}
   \centering
     \centerline{\includegraphics[width=0.86\linewidth]{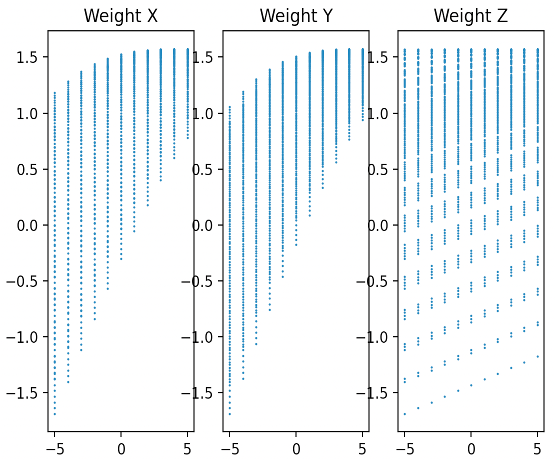}}

  \centerline{(b) $\Psi(x)=\cos(\alpha\cdot x)+x$}\medskip
 \end{minipage}
 \vspace{-0.5cm}
 \caption{Spatial distribution of activations. We illustrate the activation from X-axis, Y-axis, and Z-axis.}
 \label{fig:spatial}
 \end{figure}

\noindent
{\bf Spatial Distribution}
This part introduces the effect of the Learnable Frequency strategy. According to the Fig~\ref{fig:spatial}, the original kernel shows poor spatial inductive bias since it cannot distinguish the symmetric locations. And the Learnable Frequency enhances the spatial bias in the generated kernels.

 \begin{table}[t]
\footnotesize
    \centering
        \caption{Mirco-designs. Validate on val@SemanticKITTI.}
        \label{tab:ablation}
        \vspace{-0.3cm}
    \setlength{\tabcolsep}{0.1cm}{
    \begin{tabular}{ccc||ccc}
         \hline
	\makecell[c]{bypass\\ dilation}&mIoU&mAcc&\makecell[c]{norm \\type}&mIoU&mAcc\\ \hline 
	$1$&67.5&74.7&LayerNorm&67.5&74.7\\ 	\hline	
	$2$&66.3&72.9&BatchNorm&67.8&74.6\\
           \hline
                                       
    \end{tabular}}
    \label{tab:link_dilate_norm}
    
    \vspace{-0.3cm}
\end{table}

\subsection{More qualitative results}

\begin{figure}[tb]

\begin{minipage}[b]{.49\linewidth}
   \centering
   \centerline{\includegraphics[width=0.99\linewidth]{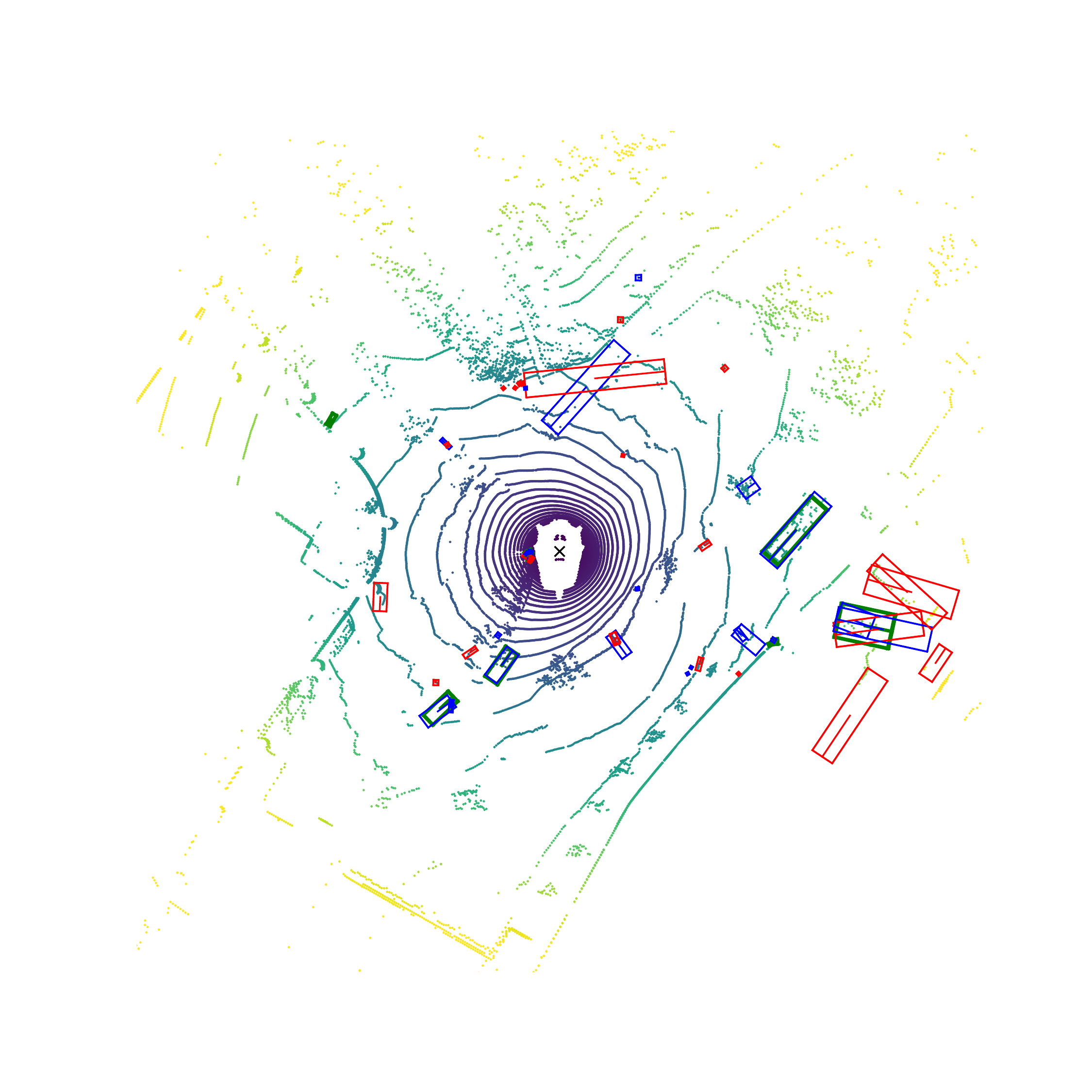}}
 %  \vspace{1.5cm}
   \centerline{(a) Baseline}\medskip
 \end{minipage}
 \hfill
 \begin{minipage}[b]{0.49\linewidth}
   \centering
  \centerline{\includegraphics[width=0.99\linewidth]{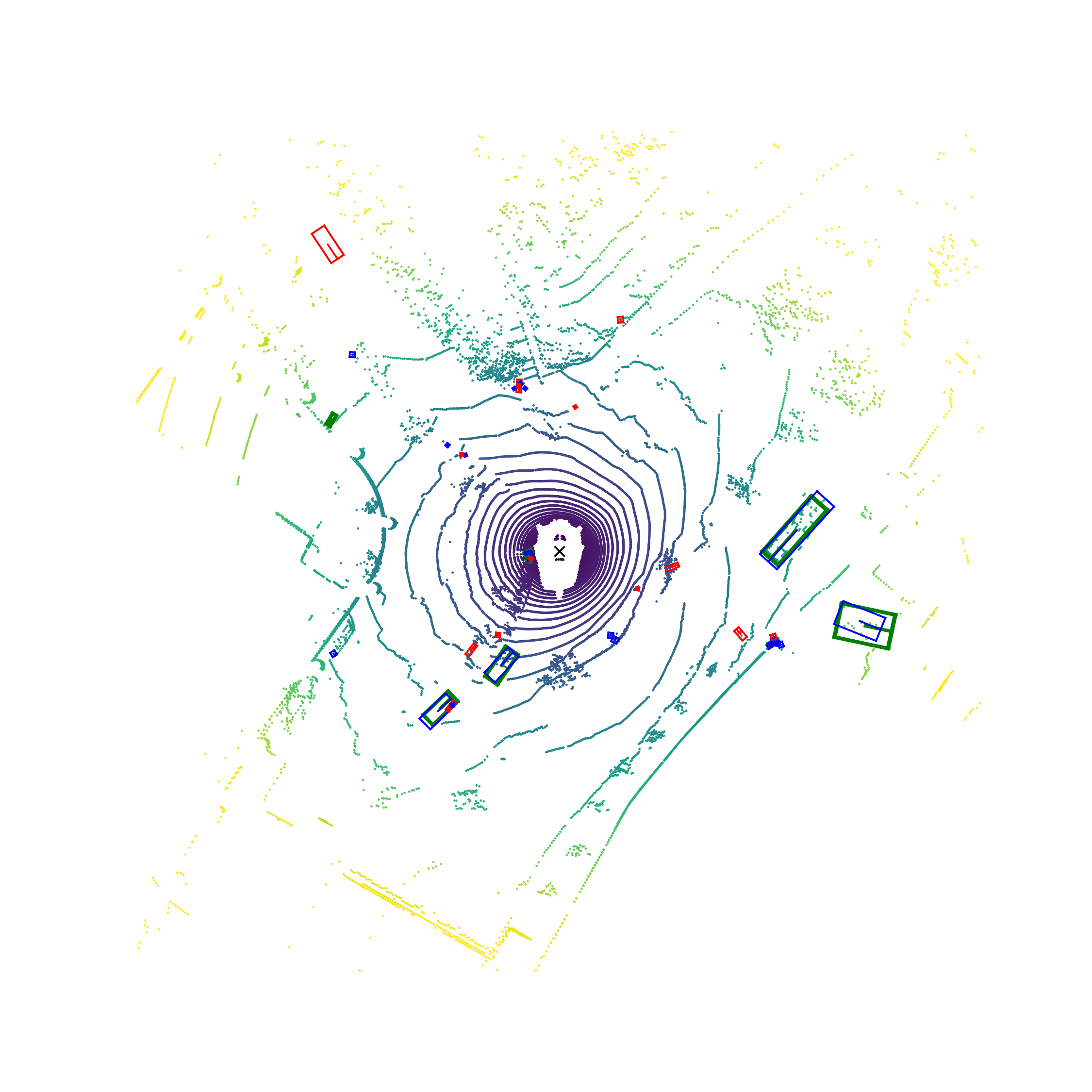}}
  \centerline{(b) LinK}\medskip
 \end{minipage}
 \caption{Detection results on nuScenes. Blue and green box are predictions and ground truth, respectively. LinK improves those remote and sparse objects in scenes. Better viewed in color and using zoom.}
 \label{fig:detnusc}
 \end{figure}

\begin{figure}[tb]
\begin{minipage}[b]{.48\linewidth}
   \centering
   \centerline{\includegraphics[width=0.8\linewidth]{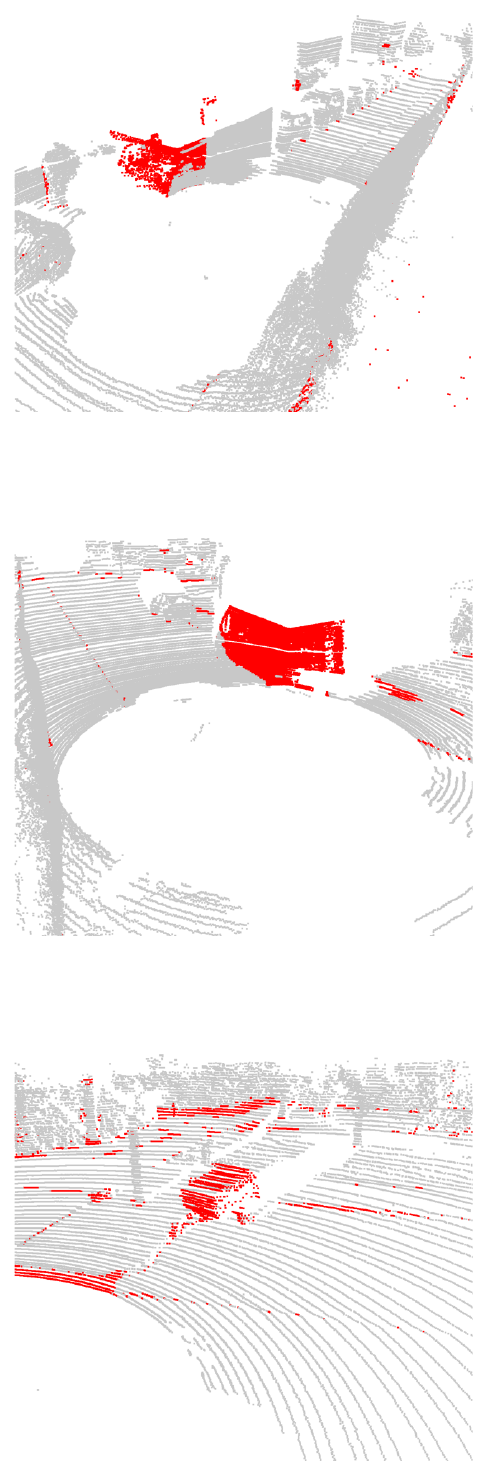}}

   \centerline{(a) Baseline}\medskip
 \end{minipage}
 \hfill
 \begin{minipage}[b]{0.48\linewidth}
   \centering
     \centerline{\includegraphics[width=0.8\linewidth]{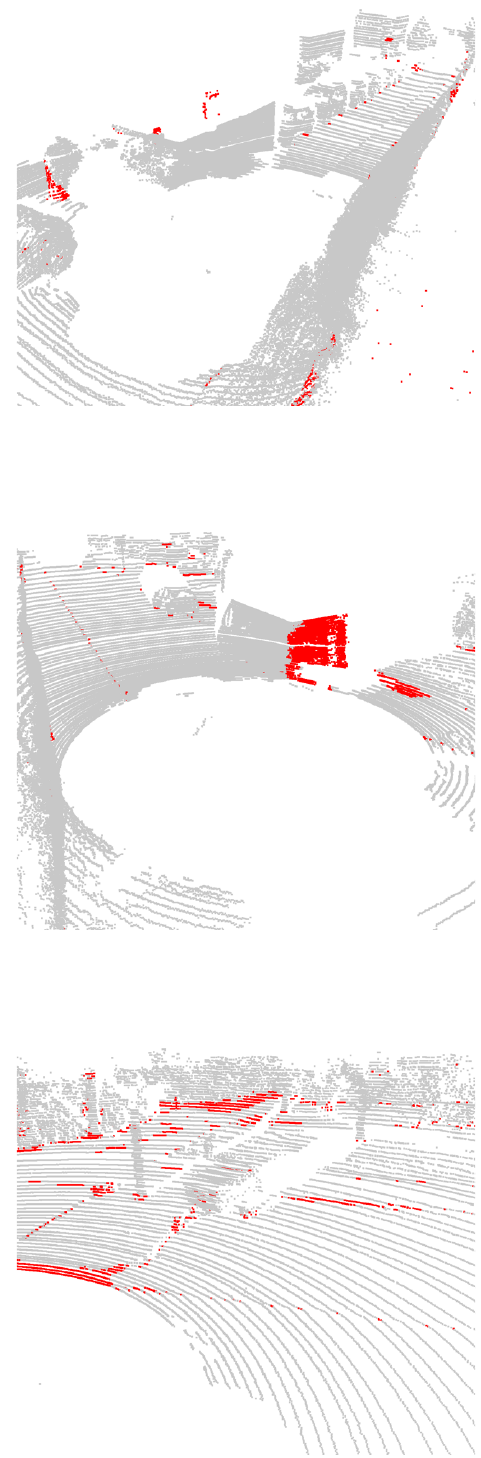}}

  \centerline{(b) LinK}\medskip
 \end{minipage}
 \vspace{-0.5cm}
 \caption{Error map of the segmentation in SemanticKITTI. Red point denote false prediction.}
 \label{fig:seg}
 \end{figure}

We provide more detection results in Fig~\ref{fig:detnusc} and more segmentation results in Fig~\ref{fig:seg} to show our improvement compared to the baseline. The sparse areas are sensitive to the noise, making the prediction unreliable. LinK improves this issue by introducing wider-range context to enhance the robustness. And the larger receptive field in LinK is more friendly to the large object.

{\small
\bibliographystyle{unsrt}
\bibliography{egbib}
}
\end{document}